\documentclass[conference]{IEEEtran}
\IEEEoverridecommandlockouts
\usepackage{cite}
\usepackage{amsmath,amssymb,amsfonts}
\usepackage{algorithmic}
\usepackage{graphicx}
\usepackage{textcomp}
\usepackage{xcolor}

\usepackage{booktabs}
\usepackage{threeparttable}
\usepackage{multirow}
\usepackage{hyperref}
\usepackage{subfigure}

\def\BibTeX{{\rm B\kern-.05em{\sc i\kern-.025em b}\kern-.08em
    T\kern-.1667em\lower.7ex\hbox{E}\kern-.125emX}}
\begin{document}

\title{Improving Attention-Based Handwritten Mathematical Expression Recognition with Scale Augmentation and Drop Attention\\
}

\author{

\IEEEauthorblockN{Zhe Li, Lianwen Jin, Songxuan Lai, Yecheng Zhu}
\IEEEauthorblockA{School of Electronic and Information Engineering\\
South China University of Technology\\
Guangzhou, China \\
Email: zheli0205@foxmail.com, eelwjin@scut.edu.cn, eesxlai@foxmail.com, claytonzyc@foxmail.com}
}

\maketitle

\begin{abstract}
Handwritten mathematical expression recognition (HMER) is an important research direction in handwriting recognition.
The performance of HMER suffers from the two-dimensional structure of mathematical expressions (MEs).
To address this issue, in this paper, we propose a high-performance HMER model with scale augmentation and drop attention.
Specifically, tackling ME with unstable scale in both horizontal and vertical directions, scale augmentation improves the performance of the model on MEs of various scales.
An attention-based encoder-decoder network is used for extracting features and generating predictions.
In addition, drop attention is proposed to further improve performance when the attention distribution of the decoder is not precise.
Compared with previous methods, our method achieves state-of-the-art performance on two public datasets of CROHME 2014 and CROHME 2016.
\end{abstract}


\section{Introduction}
Handwritten mathematical expression recognition (HMER) has been researched for more than 50 years \cite{anderson1967syntax}, and has wide applications in practice, such as human-computer interaction, office automation, and intelligent education.
As a significant branch of handwriting recognition, HMER faces many challenges, which attract researchers' interest.
Challenges of HMER include indistinguishably similar symbols, various handwriting styles, and lack of data, which are typical for handwriting recognition.
However, a mathematical expression (ME) differs from isolated characters or texts in its unique two-dimensional structure.
Therefore, HMER models should also phase the structure, which entails recognizing the spatial relations between two mathematical symbols or sub-expressions.

\begin{figure}[tbp]
\centering
\includegraphics[width=0.35\paperwidth]{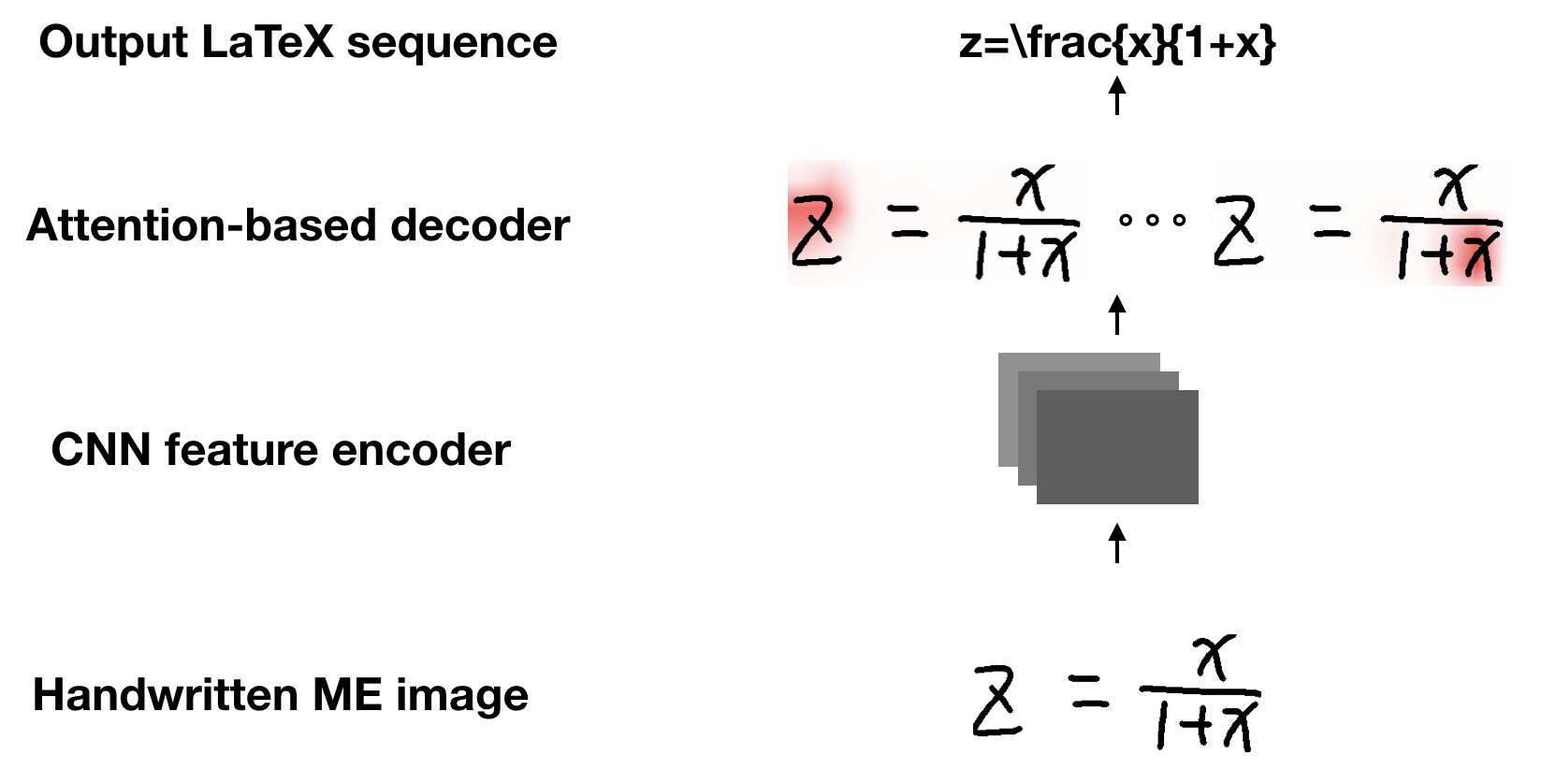}
\caption{Encoder-decoder model for HMER.}
\label{total-model}
\end{figure}

Traditional grammar-based methods, such as \cite{simistira2015recognition}, \cite{yamamoto2006line}, and \cite{alvaro2016integrated}, recognized MEs through symbol segmentation, symbol recognition, and structural analysis sequentially.
In the past few years, deep neural networks (DNNs) have made breakthrough progress in handwriting recognition \cite{xie2017learning}\cite{zhang2017drawing}.
Many DNN models for HMER were proposed by researchers, such as \cite{deng2017image}, \cite{zhang2017watch}, \cite{le2019pattern}, and \cite{wu2020handwritten}.
These studies, based on the encoder-decoder framework \cite{bahdanau2014neural}, considered HMER as an image to sequence problem, as shown in Fig. \ref{total-model}.

\begin{figure}
\centering
\subfigure[Original scale]{
\includegraphics[width=0.3\paperwidth]{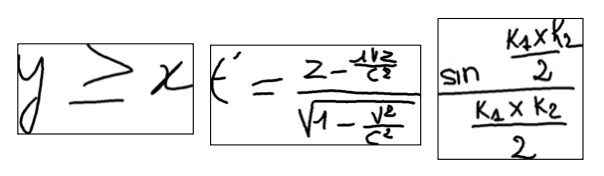}
\label{originalscale}}
\centering
\subfigure[Normalized to the fixed height]{
\includegraphics[width=0.3\paperwidth]{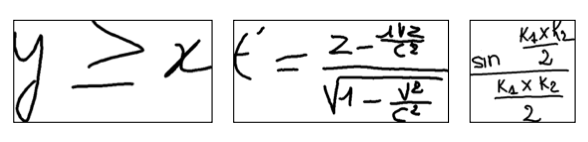}
\label{sameh}}
\subfigure[Scale augmentation]{
\includegraphics[width=0.3\paperwidth]{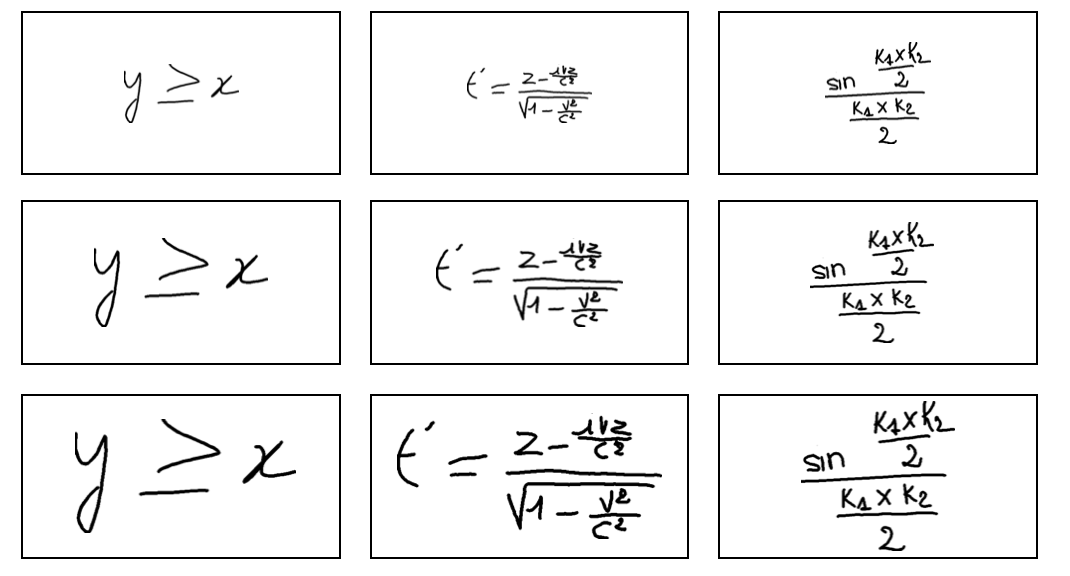}
\label{scaleaug}}
\caption{Different processing methods for ME images}
\label{datashow}
\end{figure}

Handwriting text recognition is also a typical image to sequence task, whereas in most previous works, the text are normalized to a fixed height, which is a critical process before it is fed to DNN models \cite{xie2017learning}.
However, as ME has a complex two-dimensional structure with symbols of various sizes, it is inappropriate to normalize all ME images to the same size.
As shown in Fig. \ref{originalscale}, some MEs have a multi-line structure, and the size of superscript and subscript symbol are smaller than that of other symbols.
Normalizing these ME images to the same height will degrade the DNN performance because some symbols will be too small or too large to be recognized, as shown in Fig. \ref{sameh}.
To address this issue, we propose a scale augmentation method to generate ME images for training the DNN model. 
Before being fed to DNN, MEs are augmented to another scale randomly but keeping the original aspect ratio.
Subsequently, MEs are zero-padded to a fixed size, as shown in Fig. \ref{scaleaug}.
The DNN model is then trained to generate the correct predictions from MEs of various scales.

We use an attention-based encoder-decoder framework to recognize MEs, including symbol prediction and structure phasing.
The encoder extracts features from the input image and the decoder predicts one symbol at each time step to output a sequence.  
The decoder generates an attention weight map, which represents the importance of each feature for predicting the symbol at the current time step.
When attention neglects key features, the model generates an incorrect prediction and performs worse\cite{cheng2017focusing}.
Inspired by \cite{sun2019fine}, we propose a drop attention module applied to the decoder in the training phase to alleviate this issue.
The drop attention module can assist our model to predict the correct symbol or spatial relationship when attention is imprecise through suppressing or abandoning features.
These two complementary methods, named ``scale augmentation'' and ``drop attention'', are both applied in the training phase and improve performance.

The rest of this paper is organized as follows.
In Section \ref{relw}, we briefly reviews the related works of HMER.
The proposed methods are introduced in details in Section \ref{method}.
In Section \ref{expe}, we present and analyze the experimental results.
Finally, we conclude our paper in Section \ref{conclu}.

\section{Related Works} \label{relw}

Many traditional methods for HMER were based on grammars, such as graph grammars \cite{lavirotte1998mathematical}, definite clause grammars \cite{chan2001error}, and relational grammars \cite{maclean2013new}.
Yamamoto et al. \cite{yamamoto2006line} used stochastic context-free grammars and employed a Cocke–Younger–Kasami algorithm to parse MEs.
Simistira et al. \cite{simistira2015recognition} also proposed a method based on stochastic context-free grammars and used a probabilistic support vector machine classifier to recognize spatial relations between two mathematical symbols.
Hidden Markov models have been used to recognize mathematical symbols by {\'A}lvaro et al. \cite{alvaro2014recognition} and an integrated grammar-based method has been proposed by them \cite{alvaro2016integrated}.

With the rapid development of deep learning, many methods based on the encoder-decoder framework \cite{bahdanau2014neural} have achieved excellent performance in HMER in the past few years. 
Deng et. al. \cite{deng2016you} pioneered using the encoder-decoder model in ME recognition.
Zhang et. al. \cite{zhang2017watch} further improved the performance by using a deep convolutional neural network (CNN) as an encoder and adopting coverage-based attention. 
Moreover, Zhang et. al. \cite{zhang2018multi} added an extra DenseNet \cite{huang2017densely} branch to deal with different sizes of symbols.
Zhang et. al \cite{zhang2018track} also used a recurrent neural network (RNN) as an encoder for online HMER.
Wang et. al. \cite{wang2019multi} proposed a multi-modal network with both online and offline encoder branches.
Hong et. al.  \cite{hong2019residual} used Markovian transition probability matrix for decoding.
Le et. al. \cite{le2019pattern} proposed pattern generation strategies to augment training data and improve the recognition performance.
To learn semantic-invariant features of ME, Wu et. al. \cite{wu2020handwritten} proposed a paired adversarial learning method.

\section{Proposed Methods} \label{method}
In this paper, we treat HMER as an image-to-sequence problem. 
Specifically, given an image $I$ containing an ME, our recognition model outputs the LaTeX sequence $Y$ of the ME.

\subsection{Encoder-decoder network}
\begin{figure*}[!tbp]
\centering
\includegraphics[width=0.7\paperwidth]{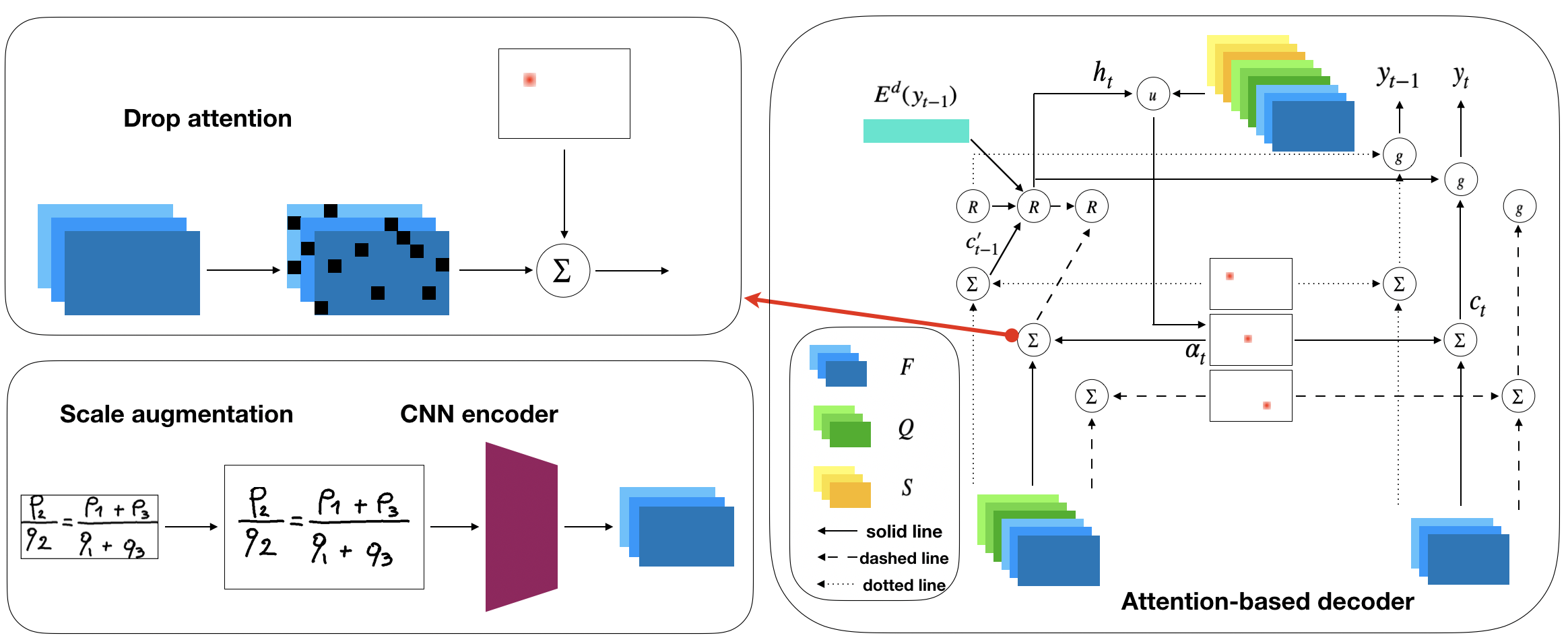}
\caption{Architecture of our proposed model. Bottom-left: an ME image after scale augmentation is fed to the CNN encoder for extracting features. Right: Attention-based decoder. $F$, $Q$, and $S$ represent CNN features, position embeddings, and coverage features, respectively. Solid, dashed, and dotted lines represent the data stream at the current, previous, and next time steps respectively. Function $u$ represents the computation of Eq. \ref{et} and Eq. \ref{alphat}. Function $R$ represents RNN. Top-left: Drop attention module applied in the training phase.}
\label{model-pic}
\end{figure*}

Our recognition model consists of a CNN encoder and an attention-based decoder, as shown in Fig. \ref{model-pic}.

We build the encoder by modifying ResNet-18 \cite{he2016deep} because CNNs have excelled in visual feature extraction.
To extract more precise features and avoid neglecting features of small-scale symbols (e.g., dot, superscript, or subscript), we set the stride of all convolutional layers in ResNet-18 to 1.
The other settings of the building block are the same as in \cite{he2016deep}.
Moreover, max pooling layers are adopted for down-sampling.
Dropout layers are also applied to alleviate network overfitting.
The detailed network configuration is shown in Tab. \ref{cnn}, where $c$, $k$, $s$, $pa$, $n$, and $p$ mean the kernel number, kernel size, stride, padding, block number, and dropout probability, respectively.

\vspace{-0.5cm}
\begin{table}[htbp]
\caption{Network configuration of the encoder}
\begin{center}
\begin{tabular}{cc}
\toprule
Layer/Block&Setting\\
\midrule
Convolution & $c=64, k=3\times3, s=1, pa=1$ \\
Max pooling & $k=2\times2, s=2$ \\
Batch norm & -\\
ReLU & -\\

Building block & c=64, n=2 \\
Max pooling & $k=2\times2, s=2$ \\
Dropout & p=0.1 \\

Building block & c=128, n=2 \\
Max pooling & $k=2\times2, s=2$ \\
Dropout & p=0.2 \\

Building block& c=256, n=2 \\
Max pooling & $k=2\times2, s=2$ \\
Dropout & p=0.3 \\

Building block & c=512, n=2 \\
Max pooling & $k=2\times2, s=2$ \\
Dropout & p=0.3 \\
\bottomrule
\end{tabular}
\label{cnn}
\end{center}
\end{table}
\vspace{-0.5cm}

An ME image $I$ is fed to CNN; subsequently, the output features $F$ are fed to the decoder.

The attention-based decoder is based on RNN and iteratively generates the target sequence $Y$ from features $F$.
At time step t, the probability of generating symbol $y_{t}$ depends on context $c_{t}$, and the current hidden state $h_{t}$ from the RNN output, as shown in Eq. \ref{probyt}:
\begin{equation}
p(y_{t}) = g(c_{t}, h_{t}),
\label{probyt}
\end{equation}
where $g(\cdot)$ is a linear function followed by a soft-max function.

We denote the size of features $F$ as $L$, where $L=H \times W$; $H$ and $W$ are the height and width of features $F$, respectively.
Context $c_{t}$ is computed as a weighted sum of features $F$, as illustrated by Eq. \ref{ct}, and $\alpha_{t, l}$ is the weight of the $l$th features of $F$ at time step t.
As the human visual mechanism does, an attention-based decoder concentrates only on a subset of features at every time step.
Trained by the back-propagation algorithm using gradient descent, the decoder can be endowed with the capability to determine which features are important for generating word $y_{t}$ at current time step t.
We choose a linear function with the activation function $\tanh$ to compute the attention weights and normalization by soft-max function in Eq. \ref{et} and Eq. \ref{alphat}, where $W^{e}$, $W^{h}$, $W^{f}$, $W^{q}$, and $W^{s}$ are trainable parameters.
Referring to \cite{wu2020handwritten}, when calculating the attention weights, position embeddings $q_{l}$ are considered to make the decoder position sensitive.
As shown in Eq. \ref{posiemb}, $E^{p,h}$ and $E^{p,v}$ are absolute position embeddings in horizontal and vertical direction, respectively.
Referring to \cite{zhang2017watch}, coverage features $s_{l}$ are adopted to alleviate over-attention and under-attention.
As shown in Eq. \ref{coverage}, $s_{l}$ are the sum of the past attention weights.

\begin{equation}
c_{t} = \sum_{l=0}^{L-1}\alpha_{t, l} \cdot f_{l},
\label{ct}
\end{equation}
\begin{equation}
e_{t, l} = W^{e} \tanh(W^{h}h_{t} + W^{f}f_{l} + W^{q}q_{l} + W^{s}s_{l}),
\label{et}
\end{equation}
\begin{equation}
\alpha_{t, l} = \frac{exp(e_{t, l})}{\sum_{i=1}^{L}exp(e_{t, i})}.
\label{alphat}
\end{equation}
\begin{equation}
q_{l}=E^{p,h}(\lfloor l / H \rfloor)+E^{p,v}(l \bmod H)
\label{posiemb}
\end{equation}
\begin{equation}
s_{l}=\sum_{\tau=0}^{t-1}\alpha_{\tau,l}
\label{coverage}
\end{equation}

The hidden state $h_{t}$ is computed by a RNN with long short-term memory (LSTM) cells.
At time step t, the RNN input is the word embedding of $y_{t-1}$ concatenated with context $c_{t-1}^{\prime}$, which is a weighted sum of not only CNN features $F$ but also position embeddings $Q$ at last time step, as illustrated by Eq. \ref{cpt}.
\begin{equation}
h_{t} = RNN([E^{d}(y_{t-1}), c_{t-1}^{\prime}], h_{t-1}),
\label{ht}
\end{equation}
\begin{equation}
c_{t}^{\prime} = \sum_{l=1}^{L}\alpha_{t, l} \cdot (f_{l} + q_{l}).
\label{cpt}
\end{equation}


\subsection{Scale augmentation}
Unlike handwritten texts, an ME has a complex two-dimensional structure and symbols of various sizes.
If multi-line MEs are normalized to the same scale though keeping the aspect ratio, some symbols (e.g., superscript, subscript, and dot) are smaller than other symbols as shown in Fig. \ref{sameh}, which increases the recognition difficulty.
Instead of normalizing MEs to the same scale, we augment MEs according to Eq. \ref{scaling},
\begin{equation}
\left\{
             \begin{array}{lr}
             x^{\prime} = kx,  &  \\
             y^{\prime} = ky, &  
             \end{array}
\right.\label{scaling}
\end{equation}
where $k$ is the scaling factor and we keep the aspect ratio constant.
In the training phase, MEs are augmented to another scale randomly and are zero-padded to the fixed size, as shown in Fig. \ref{scaleaug}.
In the test phase, MEs are zero-padded to the fixed size without augmentation.
The encoder-decoder network is trained to adapt to MEs of various scales and generate correct predictions.

\subsection{Drop attention}
As illustrated previously, attention weight $\alpha$ has a great impact on recognition performance.
When attention neglects the key features, the model will generate an incorrect prediction and perform worse.
Inspired by \cite{sun2019fine}, we propose a drop attention module in the decoder.
First, we randomly suppress the features where the corresponding attention weight $\alpha$ is the highest.
Subsequently, we randomly abandon spots on the feature maps, as in Eq. \ref{dpattn}.
\begin{equation}
f_{l}^{\prime} = \left\{
\begin{array}{cc}
             \gamma \cdot f_{l}, & {\rm if} \ l = {\rm argmax}(\alpha_{l}) \  {\rm and} \  r_{p}=0,\\
             0, & {\rm if} \ l \neq {\rm argmax}(\alpha_{l}) \  {\rm and} \  r_{s}=0,\\
             f_{l}, &otherwise,
\end{array}
\right.
\label{dpattn}
\end{equation}
where $\gamma$ is the suppress factor; $r_{p}$ and $r_{s}$ are random values and obey the Bernoulli distribution.
Features $f_{l}^{\prime}$ replace $f_{l}$ in Eq. \ref{ct} and Eq. \ref{cpt} in training phase, which assists our model to predict the correct symbol or spatial relationship when attention is imprecise.

\section{Experiments} \label{expe}
\subsection{Datasets}
The CROHME competition dataset is the largest dataset for HMER \cite{mouchere2014icfhr} \cite{mouchere2016icfhr2016}.
This dataset contains 101 classes of math symbols.
In this paper, our model was trained on CROHME 2014 training set (containing 8834 MEs) and validated on CROHME 2013 test set (containing 671 MEs).
We report the test results of expression recognition rate (ExpRate) on CROHME 2014 test set and CROHME 2016 test set, which contains 986 and 1147 MEs, respectively. 
ExpRate $\leq 1(\%)$, $\leq 2(\%)$, and $\leq 3(\%)$ denote the expression recognition rates when one, two, or three symbol-level errors are tolerable.
We convert the output LaTeX strings to label graphs and evaluate the performance with the official tools provided by the CROHME 2019 organizers \cite{mahdavi2019icdar}.
The experiment settings are consistent with the settings used by the competition participants \cite{mouchere2014icfhr} \cite{mouchere2016icfhr2016} and in most previous studies \cite{le2017training} \cite{zhang2017watch} \cite{wu2018image} \cite{le2019pattern} \cite{wu2020handwritten}.

\subsection{Implementation}
Our model was implemented in PyTorch and optimized on Nvidia TITAN Xp GPU.
The batch size was set to 8 for parallel computing.
ME images underwent scale augmentation in the training phase and were zero-padded to a fixed size $256\times1024$.
Few ME images were larger than this size before zero-padding, and they were downsized to be smaller than this size.
Scaling factor $k$ were set within $[0.5, 2]$ randomly at every training iteration.
Features were extracted by CNN from the input image, with configurations shown in Tab. \ref{cnn}, and then fed to attention-based decoder.
A drop attention module was applied in the training phase.
The suppress factor was set as 0.1. 
Bernoulli probabilities of $r_{p}$ and $r_{s}$ were 0.8 and 0.4, respectively.
We calculated the cross-entropy loss between the generated symbols and ground truth.
Adam optimizer was used to train the model, and the learning rate was 0.0001.

\subsection{Effect of scale augmentation}
We conducted comparative experiments about methods for processing ME images and without applying a drop attention module, as shown in Tab. \ref{expeaug}.
The first method involves normalizing ME images to a fixed height of 256 while keeping the aspect ratios, as shown in Fig. \ref{sameh}.
The second method is zero-padding ME images to a fixed size while the third method is applying scale augmentation before zero-padding as shown in Fig. \ref{scaleaug}.

\vspace{-0.5cm}
\begin{table}[htbp]
\caption{ExpRate(\%) of different processing methods for ME images}
\begin{center}
\begin{tabular}{ccc}
\toprule
Method & CROHME 2014 & CROHME 2016 \\
\midrule
Normalized to fixed height & 48.78 & 49.26\\
Zero-padded to fixed size & 50.00 & 47.95\\
Scale augmentation & 55.17 & 52.48\\
\bottomrule
\end{tabular}
\label{expeaug}
\end{center}
\end{table}
\vspace{-0.5cm}

As illustrated in Tab. \ref{expeaug}, the results with fixed height normalization are slightly better than that with zero-padding on CROHME 2016 but worse on CROHME 2014.
However, both these two methods achieve worse result than the proposed scale augmentation because they are weak on MEs of various scales.
The results with scale augmentation outperform zero-padding about 5\% in ExpRate, which proves that scale augmentation is effective.
The model trained with MEs using scale augmentation could extract ME discriminative features from various scales, which is important for generating a correct prediction.
Therefore, we applied scale augmentation in the subsequent experiments.

\begin{figure*}[htbp]
\centering
\includegraphics[width=0.7\paperwidth]{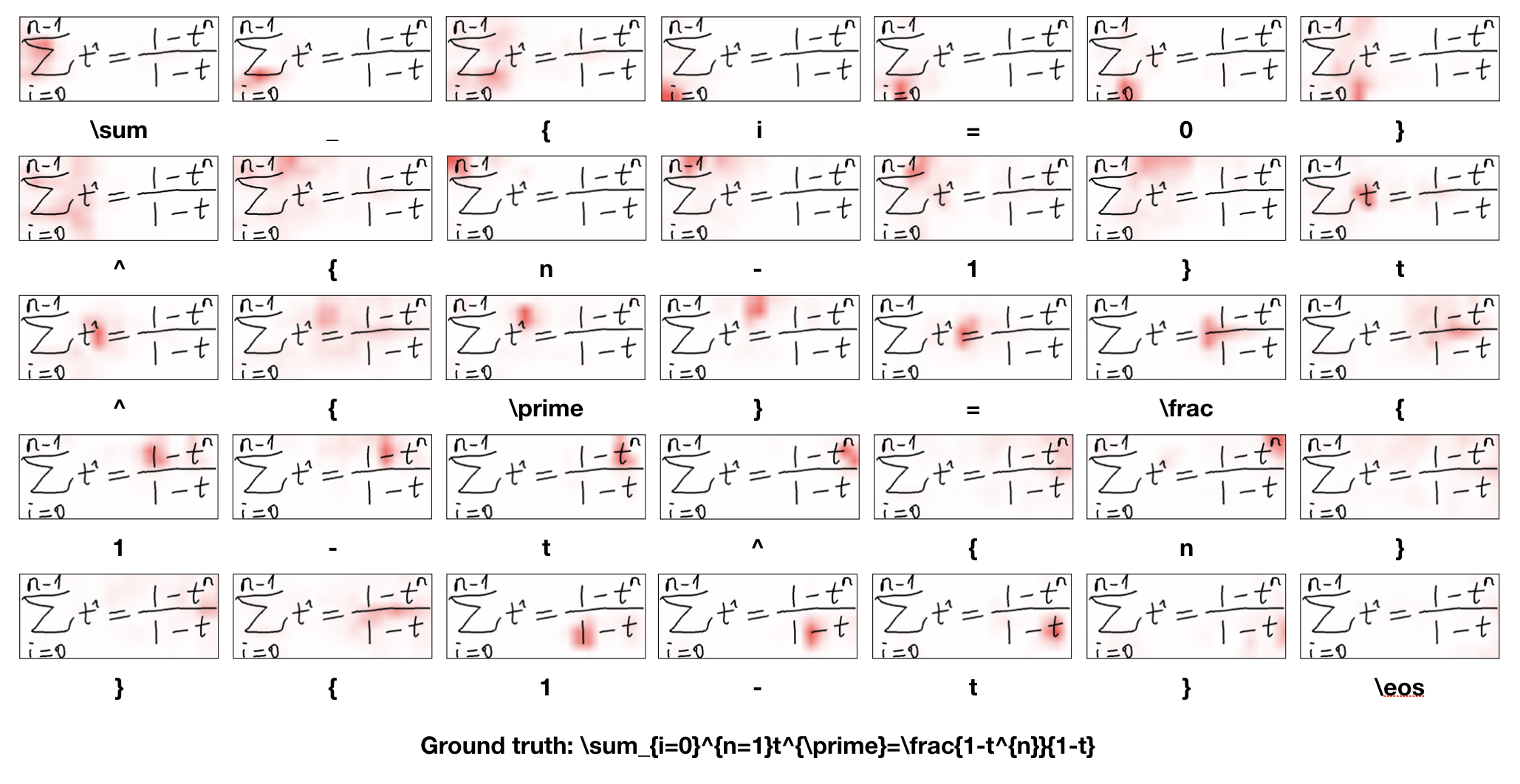}
\caption{Visualization of the attention and recognition process.}
\label{atmp}
\end{figure*}

\subsection{Effect of drop attention}
We conducted the comparative experiments to verify the effect of drop attention, and the results are shown in Tab. \ref{expedpattn}.

\vspace{-0.5cm}
\begin{table}[htbp]
\caption{ExpRate(\%) of methods with or w/o drop attention}
\begin{center}
\begin{tabular}{ccc}
\toprule
Drop attention & CROHME 2014 & CROHME 2016 \\
\midrule
$\times$ & 55.17 & 52.48\\
\checkmark & 56.59 & 54.58\\
\bottomrule
\end{tabular}
\label{expedpattn}
\end{center}
\end{table}
\vspace{-0.5cm}

As illustrated in Tab. \ref{expedpattn}, when equipped with the drop attention module, the accuracy of our model improves by 1.42\% on CROHME 2016 and 2.1\% on CROHME 2014.
The drop attention module through suppressing or abandoning some features trains model generating correct predictions when attention neglects the key features.
To further improve the performance of our model, the drop attention module was applied in the training phase of the subsequent experiments.

\subsection{Comparison with the proposed methods}

\begin{table}[!htbp]
\caption{ExpRate(\%) of different systems on CROHME 2014 test set}
\begin{center}
\begin{threeparttable}
\begin{tabular}{ccccc}
\toprule
System & ExpRate(\%) & $\leq1$(\%) & $\leq2$(\%) & $\leq3$(\%)\\
\midrule
\uppercase\expandafter{\romannumeral1} \cite{mouchere2014icfhr} & 37.22 & 44.22 & 47.26 & 50.20 \\
\uppercase\expandafter{\romannumeral2} \cite{mouchere2014icfhr} & 15.01 & 22.31 & 26.57 & 27.69 \\
\uppercase\expandafter{\romannumeral4} \cite{mouchere2014icfhr} & 18.97 & 28.19 & 32.35 & 33.37 \\
\uppercase\expandafter{\romannumeral5} \cite{mouchere2014icfhr} & 18.97 & 26.37 & 30.83 & 32.96 \\
\uppercase\expandafter{\romannumeral6} \cite{mouchere2014icfhr} & 25.66 & 33.16 & 35.90 & 37.32 \\
\uppercase\expandafter{\romannumeral7} \cite{mouchere2014icfhr} & 26.06 & 33.87 & 38.54 & 39.96 \\
WYGIWYS\tnote{*} \cite{deng2016you} & 28.70 & - & - & - \\
End-to-end \cite{le2017training} & 35.19 & - & - & - \\
WAP\tnote{*} \cite{zhang2017watch} & 44.40 & 58.40 & 62.20 & 63.10 \\
PAL \cite{wu2018image} & 39.66 & - & - & - \\
PAL\tnote{*} \cite{wu2018image} & 47.06 & - & - & - \\
DenseMSA\tnote{*}  \ \ \cite{zhang2018multi} & 52.80 & 68.10 & 72.00 & 72.70 \\
PGS \cite{le2019pattern} & 48.78 & 66.13 & 73.94 & 79.01 \\
PAL-v2 \cite{wu2020handwritten} & 48.88 & 64.50 & 69.78 & 73.83 \\
PAL-v2\tnote{*} \cite{wu2020handwritten} & 54.87 & 70.69 & 75.76 & 79.01 \\
ours & 56.59 & 69.07 & 75.25 & 78.60\\
ours\tnote{*} & \textbf{60.45} & \textbf{73.43} & \textbf{77.69} & \textbf{80.12} \\
\bottomrule
\end{tabular}
\begin{tablenotes}
\item{*} Utilizing an ensemble of 5 differently initialized recognition models.
\end{tablenotes}
\label{crohme14}
\end{threeparttable}
\end{center}
\end{table}

Tab. \ref{crohme14} compares the results of our model with the systems submitted at CROHME 2014 and recent offline attention-based HMER models.
Systems \uppercase\expandafter{\romannumeral1} to \uppercase\expandafter{\romannumeral7} were participating systems except for system \uppercase\expandafter{\romannumeral3} because of using private training data.
We do not present results from \cite{zhang2018track}, \cite{wang2019multi} and \cite{hong2019residual} because their models were trained on handwriting trajectory data. 
A trajectory provides handwriting order information, which is useful for distinguishing visually similar symbols (e.g., ``$\alpha$'' and ``$a$'').
The attention-based models listed in Tab. \ref{crohme14} are all trained on offline ME images with only LaTeX level labels, including our method.
The sign ``$*$'' in Tab. \ref{crohme14} denotes using an ensemble of five differently initialized recognition models to improve the performance \cite{zhang2017watch}.
In the case of with and without ensemble, our method outperforms the state-of-the-art model PAL-v2 by 5.58\% and 7.71\%, respectively.

\begin{table}[!htbp]
\caption{ExpRate(\%) of different systems on CROHME 2016 test set}
\begin{center}
\begin{threeparttable}
\begin{tabular}{ccccc}
\toprule
System & ExpRate(\%) & $\leq1$(\%) & $\leq2$(\%) & $\leq3$(\%)\\
\midrule
Wiris \cite{mouchere2016icfhr2016} & 49.61 & 60.42 & 64.69 & - \\
Tokyo \cite{mouchere2016icfhr2016} & 43.94 & 50.91 & 53.70 & - \\
Sao Paolo \cite{mouchere2016icfhr2016} & 33.39 & 43.50 & 49.17 & - \\
Nantes \cite{mouchere2016icfhr2016} & 13.34 & 21.02 & 28.33 & - \\
WAP\tnote{*} \cite{zhang2017watch} & 44.55 & 57.10 & 61.55 & 62.34 \\
DenseMSA\tnote{*}  \ \ \cite{zhang2018multi} & 50.10 & 63.80 & 67.40 & 68.50 \\
PGS \cite{le2019pattern} & 45.60 & 62.25 & 70.44 & 75.76 \\
PAL-v2 \cite{wu2020handwritten} & 49.61 & 64.08 & 70.27 & 73.50 \\
PAL-v2\tnote{*} \cite{wu2020handwritten} & 57.89 & 70.44 & \textbf{76.29} & \textbf{79.16} \\
ours & 54.58 & 69.31 & 73.76 & 76.02\\
ours\tnote{*} & \textbf{58.06} & \textbf{71.67} & 75.59 & 77.59 \\
\bottomrule
\end{tabular}
\begin{tablenotes}
\item{*} Utilizing an ensemble of 5 differently initialized recognition models.
\end{tablenotes}
\label{crohme16}
\end{threeparttable}
\end{center}
\end{table}

Tab. \ref{crohme16} compares the results of our model with the systems submitted at CROHME 2016 and recent offline attention-based HMER models.
Our method outperforms ``Wiris'' system with a large margin, which won the first place in CROHME 2016 using only the official handwritten MEs training data.
Moreover, our method also performs better than the state-of-the-art model PAL-v2 by 4.97\% in the case of without ensemble and slightly better than with ensemble by 0.17\%.

In general, the results have shown that our method has excellent performance and potential to achieve higher performance.

\subsection{Visualization}\label{visula}
Fig. \ref{atmp} illustrates the attention-based decoding process.
Attention weights are visualized in red. 
Darker red denotes a higher weight in the attention map.
At each time step, features from the CNN encoder are summed according to the attention weight.
The model outputs one prediction symbol and composes a LaTex string gradually until the output symbol is ``$\backslash$eos'', which means ``end of string.''
Thus, our model can correctly predict symbols of various sizes and the relationships between them.
The decoder can focus on discriminative features of symbols and be ``distracted'' when outputting grammar symbols, such as ``\{''.
Because we used RNN network to compose our decoder, there is an implicit language model learned in training phase.
Therefore, when generating symbol ``\_'' or ``$\wedge$'' at the last time step, our model can generate ``\{'' without referring to the attention features.

\section{Conclusion} \label{conclu}
In this paper, we propose a scale augmentation method to address the recognition problem of MEs of various scales, which is caused by the complex two-dimensional structure of ME.
Moreover, a new drop attention module is proposed to further improve performance by training the model to generate predictions when attention is imprecise.
The experimental results indicated that these two technologies assist our method achieving state-of-the-art performance, compared with the previous methods without an ensemble.
In future work, we plan to extend our method to deal with online HMER.

\section*{Acknowledgement}

This research is supported in part by NSFC (Grant No.: 61936003),  GD-NSF (no.2017A030312006), the National Key Research and Development Program of China (No. 2016YFB1001405), Guangdong Intellectual Property Office Project (2018-10-1), and Fundamental Research Funds for the Central Universities (x2dxD2190570).


\bibliography{ref}
\bibliographystyle{ieeetr}

\end{document}